\documentclass[11pt,authoryear]{elsarticle}
\usepackage[T1]{fontenc}
\usepackage{lmodern,amsmath,amssymb,booktabs,graphicx}
\usepackage[hidelinks]{hyperref}
\newcommand{\system}{\textsc{SteelAgent}}
\newcommand{\papertitle}{SteelAgent: LLM-Orchestrated, Physics-Informed Steel Property Prediction with Audited Generalization and Calibrated Uncertainty}
\hypersetup{pdftitle={\papertitle},pdfauthor={Aleksandr Volkov, Roman Sultimov, Mikhail Kuzin, Yury Maximov},pdfsubject={Expanded preprint. A shorter version was published in the IJCAI-26 Demonstrations Track, pp. 8547-8550 (2026). Earlier publication DOI: 10.24963/ijcai.2026/1000.}}
\makeatletter
\def\ps@pprintTitle{\let\@oddhead\@empty\let\@evenhead\@empty\def\@oddfoot{\hfil\thepage\hfil}\let\@evenfoot\@oddfoot}
\makeatother
\begin{document}
\begin{frontmatter}
\title{\papertitle\tnoteref{publication}}
\tnotetext[publication]{A shorter version of this work, \emph{SteelAgent: An LLM-Orchestrated System for Physics-Informed Steel Property Prediction and Generalization Auditing}, was published in the Proceedings of the Thirty-Fifth International Joint Conference on Artificial Intelligence (IJCAI-26), Demonstrations Track, pp.~8547--8550 (2026). \href{https://doi.org/10.24963/ijcai.2026/1000}{doi:10.24963/ijcai.2026/1000}.}
\author{Aleksandr Volkov}
\author{Roman Sultimov}
\author{Mikhail Kuzin}
\author{Yury Maximov}
\begin{abstract}
Steel property models can perform well on random test splits even when they extrapolate poorly to unfamiliar alloys. SteelAgent couples metallurgical descriptors and machine-learning models with a language-model interface for property prediction, uncertainty reporting and inverse design. We examine how the choice of held-out unit changes the conclusions drawn from SteelBench tensile-strength evaluations. For gradient boosting, pooled out-of-fold $R^2$ fell from 0.859 on a random split to 0.113 when steel families were held out. Empirical coverage of nominal 90\% prediction intervals ranged from 91.0\% under random splitting to 37.6\% on held-out sources. The effect of physics-derived features depended on the evaluation. They increased gradient-boosting $R^2$ by 0.2534 in a 17-family holdout analysis and decreased it by 0.358 in a separate four-grade laboratory analysis. We report group-level errors and a synthetic regression experiment illustrating how calibration can deteriorate on held-out groups. These results motivate reporting the evaluation population alongside a property estimate or compliance score.
\end{abstract}
\end{frontmatter}

\section{Introduction}
When an engineer requests a tensile-strength estimate for an alloy, the most relevant test set depends on what is already known about that alloy. Predicting another heat of a familiar grade, predicting an unfamiliar grade, and transferring a model to a new measurement source pose different questions. The distinction matters when the estimate guides alloy selection or prompts an engineer to order a qualification test.

Prior work has used machine learning to predict steel properties, design alloys and organize materials data \citep{Xiong2020,Wen2019,Choudhary2022}. Train--test overlap and choices about data partitioning can make performance difficult to interpret \citep{Kapoor2023}. In a random split, related grades and measurements from the same source may occur on both sides. An evaluation intended to establish transfer across grades or sources must explicitly withhold those units.

The IJCAI demonstration introduced \system{}, an interactive system for querying metallurgical models and checking specifications \citep{Volkov2026demo}. Here we bring together the benchmark evaluations behind the demonstration, distinguish pooled and group-level performance, and compare two ablations of physics-derived features that used different populations. We also examine what interval coverage can tell a user about a change in operating conditions. The steel results are a secondary analysis of the SteelBench evaluation tables \citep{SteelBenchManuscript}. A separate, reproducible synthetic experiment examines group holdout for a ridge model.

\section{The system and the prediction task}
\system{} connects three components. A physics layer derives metallurgical descriptors from composition and heat treatment. A prediction layer supplies model estimates, model comparisons and uncertainty intervals. A language model selects tools and explains their outputs. Matbench and matminer provide benchmarks and feature-processing tools for materials modeling \citep{Dunn2020,Ward2018}. Coscientist and ChemCrow demonstrate complementary uses of language models to coordinate scientific tools \citep{Boiko2023,Bran2024}.

The demonstration describes six operations: looking up a reference grade, computing physical descriptors, predicting a property, checking a specification, comparing models and screening candidate compositions for inverse design. For example, a query about quenched and tempered AISI~4340 first resolves the grade and treatment, then computes inputs for a tensile-strength model. The demonstration reports a prediction of 1,746~MPa and a nominal 90\% interval of [1,546, 1,945]~MPa for its example \citep{Volkov2026demo}. Whether such an interval is useful depends on how well it covers measurements in the intended deployment population.

The descriptors include martensite start temperature, carbon equivalent and ideal critical diameter. Their physical interpretation draws on established heat-treatment relations \citep{Andrews1965,Grossmann1964}. For example, with composition expressed as weight percentages, the carbon equivalent used in the demonstration combines the alloying elements as
\[
  \mathrm{CE}_{\mathrm{IIW}} =
  \mathrm{C}+\frac{\mathrm{Mn}}{6}
  +\frac{\mathrm{Cr}+\mathrm{Mo}+\mathrm{V}}{5}
  +\frac{\mathrm{Ni}+\mathrm{Cu}}{15}.
\]
A practical implementation must specify how it handles missing elements before evaluating this expression. Recording the composition, the rule used and its output lets an engineer audit a calculation separately from the language model's interpretation.

\section{Data, evaluation units and metrics}
The broader system inventory described in the demonstration contains 3,741 heat-treatment records, 1,234 grades and 27 sources. The benchmark analyses use different subsets of that inventory. The 1,636-record Silver core spans 594 grades. The SteelBench v1.0 dataset record describes 17 families \citep{SteelBench}. A separate extended Silver compilation contains 2,219 records and supplies the learning-curve and seed-sensitivity analyses. The laboratory Gold tier comprises 68 heats, with 60 mechanical tests across four grades used in the grade-holdout ablation \citep{SteelBenchManuscript}. The public files are an open subset of the benchmark materials.

The headline outcome is ultimate tensile strength in MPa. We use the reported pooled out-of-fold coefficient of determination, $R^2=1-\sum_i(y_i-\hat y_i)^2/\sum_i(y_i-\bar y)^2$, and root mean squared error (RMSE). Each prediction in the pooled score comes from a fit that withheld its test fold. A score calculated within one held-out family uses that family's own response mean and denominator. The pooled score can therefore be positive even when many within-family scores are negative.

Four evaluation protocols probe different forms of transfer. Random $k$-fold assigns individual records to folds. Group $k$-fold withholds all records for a grade together. Leave-one-family-out (LOFO) withholds a metallurgical family. Leave-one-source-out (LOSO) withholds a provenance group. These protocols change group composition and sometimes sample size. They do not isolate a unique physical cause of the performance differences.

\section{Generalization across families and sources}
Table~\ref{tab:main} gives the SteelBench point-prediction results for the benchmark's gradient-boosting regressor (GBR) and CatBoost. Random-split $R^2$ exceeded 0.85 for both models. When complete families were held out, GBR fell to 0.1134 and CatBoost to 0.3544. Holding out sources produced the largest RMSE for both models, although the lowest GBR $R^2$ occurred under family holdout. RMSE and $R^2$ need not order the protocols identically because the spread of observed tensile strengths also changes.

\begin{table}[htbp]
\centering\small
\caption{Pooled out-of-fold tensile-strength results from the SteelBench evaluation tables (seed 42). RMSE is in MPa. The headline family holdout uses a filtered set of 1,351 records.}\label{tab:main}
\begin{tabular}{lrrrrr}
\toprule
& & \multicolumn{2}{c}{GBR} & \multicolumn{2}{c}{CatBoost}\\
\cmidrule(lr){3-4}\cmidrule(lr){5-6}
Holdout & $n$ & $R^2$ & RMSE & $R^2$ & RMSE \\
\midrule
Random & 1,636 & 0.8591 & 114.1 & 0.8608 & 113.4\\
Grade group & 1,636 & 0.6685 & 175.0 & 0.7049 & 165.1\\
Family & 1,351 & 0.1134 & 218.3 & 0.3544 & 186.3\\
Source & 1,636 & 0.1691 & 277.0 & 0.1060 & 287.3\\
\bottomrule
\end{tabular}
\end{table}

\paragraph{Differences within held-out groups.}
CatBoost produced a negative within-family $R^2$ in eight of the twelve reported families, including low-carbon steel ($-2.925$) and the remaining low-alloy group ($-0.018$). Four scores were positive: carbon steel (0.117), normalized low-alloy steel (0.147), duplex stainless (0.345) and martensitic stainless (0.489). A negative score means that the sum of squared errors exceeded the sum obtained by using the held-out group's mean, a retrospective reference value. The five reported within-source GBR scores also varied: four were negative and the Kaggle measured group was slightly positive (0.060). The lowest score was in the laboratory group ($-1.273$).

\paragraph{Learning curves and seed sensitivity.}
In the five-seed learning-curve study, the gap between random-split and grade-grouped GBR $R^2$ was 0.341 at the 20\% setting, 0.317 at 50\% and 0.217 at 100\%. The nominal fractions corresponded to different numbers of records under the two protocols; at 100\%, the tabulated counts were 2,219 and 1,209, respectively. The gap persisted over the measured settings. These results do not establish how it would behave with a larger matched training set. In a separate 20-seed study, the mean (standard deviation) was 0.7245 (0.0051) on random splitting and 0.4551 (0.0296) on grade grouping. Training on extended Silver and testing on 60 Gold measurements yielded negative pooled $R^2$ for Ridge ($-3.419$), GBR ($-5.277$) and CatBoost ($-3.942$).

\section{When do physics-derived features help?}
The demonstration reported an increase of about 0.25 in family-holdout GBR $R^2$. The full-family ablation supports that result: adding descriptors moved $R^2$ from $-0.0336$ to 0.2198, an increase of 0.2534 (Table~\ref{tab:ablations}). This experiment included all 17 families, without the minimum-family-size filter used for the headline family result in Table~\ref{tab:main}. CatBoost's score declined in the same ablation.

\begin{table}[htbp]
\centering\small
\caption{Tensile-strength ablations from SteelBench. Changes are differences in pooled $R^2$ within each row. The Silver and Gold tests use different holdout units and descriptor sets.}\label{tab:ablations}
\begin{tabular}{llrrr}
\toprule
Population and holdout & Model & Baseline & Augmented & Change\\
\midrule
Silver, family & GBR & $-0.0336$ & 0.2198 & $+0.2534$\\
Silver, family & CatBoost & 0.2400 & 0.1047 & $-0.1353$\\
\addlinespace
Gold, grade & Ridge & $-2.042$ & 0.199 & $+2.241$\\
Gold, grade & GBR & $-1.385$ & $-1.743$ & $-0.358$\\
Gold, grade & CatBoost & $-1.316$ & $-1.255$ & $+0.061$\\
\bottomrule
\end{tabular}
\end{table}

The Gold experiment held out each of four grades in turn. It used 60 laboratory tests and a separate set of physics descriptors. Feature augmentation improved the ridge score by 2.241, decreased the GBR score by 0.358, and improved the CatBoost score by 0.061. Even after augmentation, both tree models had negative $R^2$ in this small test. For a prospective prediction, the user needs to know which grade or family is new, which descriptor set is available, and how the chosen model performed under that particular holdout.

\section{Prediction intervals under distribution shift}
The benchmark used conformalized quantile regression (CQR) with quantile-loss gradient boosting \citep{Romano2019}. Split conformal methods provide marginal coverage under exchangeability of calibration and test observations \citep{Vovk2005,Angelopoulos2023}. Holding out an entire source can change the residual distribution, so nominal coverage under exchangeability need not carry over unchanged \citep{Tibshirani2019}.

\begin{table}[htbp]
\centering\small
\caption{Empirical tensile-strength interval coverage at 90\% nominal coverage. Width is the mean interval width. Sample counts are specific to this evaluation.}\label{tab:coverage}
\begin{tabular}{lrrrr}
\toprule
Holdout & $n$ & Point-model $R^2$ & Coverage & Width (MPa)\\
\midrule
Random & 1,636 & 0.846 & 91.0\% & 378\\
Grade group & 1,636 & 0.705 & 83.1\% & 357\\
Family & 1,357 & $-0.054$ & 76.1\% & 385\\
Source & 1,636 & 0.125 & 37.6\% & 412\\
\bottomrule
\end{tabular}
\end{table}

Mean widths varied from 357 to 412~MPa while realized coverage ranged from 37.6\% to 91.0\% (Table~\ref{tab:coverage}). Under source holdout the shortfall relative to nominal coverage was 52.4 percentage points. For a new source, a wide interval by itself offers little assurance that calibration remains appropriate. Once measured outcomes arrive, a rolling coverage estimate can flag a shortfall for investigation. An 85\% trigger is one possible operational threshold. These tables do not establish its false-alarm rate or identify the type of shift from coverage alone.

\paragraph{From an interval to a compliance score.}
A nominal 90\% CQR interval does not identify the conditional probability that a particular heat exceeds a specification minimum $y_{\min}$. The demonstration offers a separate Gaussian approximation. For a symmetric interval with half-width $h$ and point estimate $\hat y$, it sets $\hat\sigma=h/\Phi^{-1}(0.95)$ and computes $\Phi((\hat y-y_{\min})/\hat\sigma)$, where $\Phi$ is the standard normal distribution function. The calculation assumes an approximately centered Gaussian conditional error distribution. CQR intervals can be asymmetric, and marginal calibration does not establish that assumption. Validation against observed pass/fail outcomes is needed before using this score for qualification.

\section{Synthetic group-holdout experiment}
We also tested coverage in a regression problem with a known data-generating process. Each of 1,636 observations belongs to one of twelve groups. The eleven-dimensional covariate vector is the sum of a group center drawn from $N(0,4I)$ and independent $N(0,0.64I)$ noise. The response combines a linear term, an interaction, a squared feature, a random group intercept and independent Gaussian noise. We fit ridge regression with penalty one and calibrate symmetric intervals on absolute residuals from a separate calibration set. The ancillary script specifies the generator, splits, quantile convention and seeds (1000--1019).

In the first setting, we randomly held out individual observations. In the second, we held out two randomly selected groups. In the third, we held out two groups chosen by the group-center projection rule in the script and added a nonlinear term to the response function for that setting. Mean empirical coverage (standard deviation across seeds) was 90.6\% (2.0 percentage points), 54.3\% (28.5 points) and 52.1\% (23.8 points), respectively. The last setting changes the response function as well as the choice of test groups, so the difference from the second setting cannot be attributed to group selection alone.

This construction illustrates one way coverage can deteriorate. A change in the input distribution alone does not imply undercoverage. For example, if a fixed predictor has an absolute error of one for every input, a calibrated interval with half-width one continues to cover even when the input distribution moves to an entirely different region. The experiment supplements the benchmark evidence without establishing a universal relation between group shift and coverage.

\section{Discussion and limitations}
The choice of evaluation unit changes what a steel-property score permits an engineer to conclude. The Silver results suggest that a random-split evaluation can substantially overstate accuracy on unseen families. The two physics-feature ablations also show why the description of the held-out unit must travel with the reported result. A gain in a full-family analysis and a loss in a four-grade laboratory test answer different engineering questions.

The analysis uses the SteelBench tabulations for steel outcomes and a separate rerunnable simulation for sensitivity to group structure. The reported protocols differ in population, group sizes and, for some comparisons, features. Original per-heat predictions would be needed to recompute uncertainty in the steel metrics or to compare models on identical held-out units. The system demonstrations cover tensile and yield properties. The quantitative comparisons here concern tensile strength. We have not validated the compliance score in prospective alloy qualification or assessed the language-model interface in an expert-user study.

\section*{Data and code availability}
The public SteelBench v1.0 record provides the open data files and release documentation \citep{SteelBench}. The broader system inventory is described in the IJCAI demonstration \citep{Volkov2026demo}. The point-prediction, feature-ablation and coverage tables are attributed to the SteelBench benchmark manuscript \citep{SteelBenchManuscript}. Restricted source measurements and historical per-heat prediction logs are outside the scope of this release. The synthetic experiment can be reproduced with the ancillary file \texttt{synthetic\_coverage.py} (Python~3.14.8, NumPy~2.5.4).
\section*{Funding}
This work received support from the Ministry of Economic Development of the Russian Federation (grant No.~139-15-2025-012).
\section*{Author contributions}
Aleksandr Volkov contributed to conceptualization, software, investigation and writing the original draft. Roman Sultimov contributed to methodology, software, validation and writing the original draft. Mikhail Kuzin contributed to data curation and investigation. Yury Maximov contributed to conceptualization, supervision, and manuscript review and editing.
\section*{Declaration of competing interest}
The authors declare no known competing financial interests or personal relationships that could have appeared to influence this work.
\bibliographystyle{elsarticle-harv}
\bibliography{refs}
\end{document}